# Attention-Enhanced Lightweight Hourglass Network for Human Pose Estimation

**Marsha Mariya Kappan[1], Eduardo Benitez Sandoval[2], Erik Meijering[1], Francisco Cruz[1,3]**
[1]School of Computer Science and Engineering, University of New South Wales, Sydney, Australia
[2]School of Art and Design, Creative Robotics Lab, University of New South Wales, Sydney, Australia
[3]Escuela de Ingeniería, Universidad Central de Chile, Santiago, Chile
Emails: {m.marsha_mariya_kappan, e.sandoval, erik.meijering, f.cruz}@unsw.edu.au

## Abstract

Pose estimation is a critical task in computer vision with a wide range of applications from activity monitoring to human-robot interaction. However, most of the existing methods are computationally expensive or have complex architecture. Here we propose a lightweight attention based pose estimation network that utilizes depthwise separable convolution and Convolutional Block Attention Module on an hourglass backbone. The network significantly reduces the computational complexity (floating point operations) and the model size (number of parameters) containing only about 10% of parameters of original eight stack Hourglass network. Experiments were conducted on COCO and MPII datasets using a two stack hourglass backbone. The results showed that our model performs well in comparison to six other lightweight pose estimation models with an average precision of 72.07. The model achieves this performance with only 2.3M parameters and 3.7G FLOPs.

## 1 Introduction

Human pose estimation (HPE) is a method of identifying locations and orientations of body joints from videos or images. It is of significant importance to a large range of applications such as action recognition [Chéron *et al.*, 2015], human computer interaction [Cheng *et al.*, 2021], sports [Ren, 2023], healthcare [Romeo *et al.*, 2023], etc. The field of human pose estimation improved a lot after the emergence of convolutional neural networks (CNNs) for keypoint detection. Samkari et al. [2023] carried out a detailed study about HPE using deep learning. Their systematic literature review provides an overview of the evolution of pose estimation and about the state-of-the-art methods in this area. The review summarizes the limitations and advantages of various pose estimation approaches and their accuracy. To achieve higher accuracy, many among the existing methods use very deep and wide neural network architectures since these networks can capture a wide range of dependencies and represent complex patterns efficiently.

The massive adoption of deep learning has greatly improved the HPE methods by allowing them to learn robust features from large datasets. Some methods such as DeepPose [Toshev and Szegedy, 2014] treat HPE as a regression problem by directly taking image input and predicting the coordinates of the joints. This approach is computationally effective as it makes the prediction simpler by predicting the numerical values of coordinates. However, the downside is that they perform worse than detection-based methods [Sun *et al.*, 2017]. Also, as they predict keypoint coordinates directly without explicitly considering the spatial relationships between body joints they suffer from incorrect predictions in case of complex scenarios. On the other hand, methods such as hourglass proposed by [Newell *et al.*, 2016] uses detection-based approaches. These models output heatmaps (using the Gaussian distribution to represent the likelihood of the presence of a keypoint) for each joint. The detection-based methods are more accurate as they consider the spatial context too, which facilitates the localization of keypoints. However, the computational demands of such models are comparatively higher. They need a large number of parameters, which increases the model size and memory demands, and it also significantly increases the number of floating point operations (FLOPs), which increases processing power demands. Thus, this might affect deploying such methods in resource constrained devices such as robots. Therefore, it is important to develop a network by considering these factors.

In this paper we propose a lightweight human pose estimation network referred to as LAP (**L**ightweight **A**ttention based **P**ose estimation network) based on an hourglass architecture and attention mechanism. The aim of LAP is to reduce the computational complexity and the number of parameters of the network. The original architecture of hourglass is redesigned and a Convolutional Block Attention Module (CBAM) proposed by [Woo *et al.*, 2018] is introduced to LAP to focus on the most important features in both channel and spatial dimensions.

The contributions of this work are summarized as follows:

- A lightweight attention based pose estimation network (LAP) is proposed which employs an hourglass backbone and attention mechanism to balance good accuracy with reduced complexity.

- The network replaces the standard convolution layer with a depthwise separable convolution layer to reduce the model size and computational cost.

- A CBAM attention module is integrated into the original hourglass network to improve the model's ability to focus on relevant spatial and channel features.

- LAP replaces the traditional Rectified Linear Unit (ReLU) activation function with the Exponential Linear Unit (ELU) for better handling of the vanishing gradient problem and enhancing the learning capability of the network.

## 2 Related Works

In recent years deep learning has revolutionized the field of HPE due to its ability to learn and extract relevant features instead of traditional methods which highly relied on handcrafted features. The Stacked Hourglass network [Newell *et al.*, 2016] uses a stack of eight symmetric encoder decoder architectures to capture features at multiple scales through a series of downsampling and upsampling operations. This method which uses heatmap generation for keypoint extraction obtained good results on benchmark datasets like MPII [Andriluka *et al.*, 2014] by efficiently modelling spatial relationships between keypoints. Other popular methods that depend on downsampling and upsampling operations are cascaded pyramid networks [Chen *et al.*, 2018] and Simple Baseline [Xiao *et al.*, 2018]. Later Sun et al. [2019] proposed HRNet (High Resolution Network) which maintains high resolution representations throughout the complete network rather than recovering high resolution representations from low resolution representations. However, the need to maintain the high resolution representations throughout the entire network increases the computational demand making it challenging to deploy in resource constrained devices.

As deep neural networks (DNN) obtained advancements in various tasks the complexity of the models also grew making their deployment in resource constrained devices challenging. To address these challenges, researchers developed lightweight models which balance good accuracy and low computational cost. Lightweight networks such as Xception [Chollet, 2017] reduce complexity using techniques like depthwise separable convolution, which splits standard convolutions to simpler lightweight operations. This results in reduced computational costs. ShuffleNet [Zhang *et al.*, 2017] is another lightweight network which uses pointwise group convolution and channel shuffling which helps to reduce parameters and floating point operations while maintaining good accuracy and speed.

In addition to these techniques, methods such as quantization and pruning also help to reduce storage requirements and speed up the computations. Quantization compresses the model by reducing the precision of weights and activations which in turn reduces computational load. Wu et al. [2016] proposed Quantized CNN (Q-CNN), a method which involved quantization of convolutional layers and fully connected layers. The Q-CNN framework showed improvements in terms of both speed and memory. Pruning helps to create a faster network without compromising performance by eliminating redundant parameters. Han et al. [2015] proposed a deep compression method which identifies and removes the unimportant connections in the neural network. The removal of these excess connections resolves the need for large storage and computational power. The process starts with normal network training. Instead of learning final weight values, though, the network learn about important connections and then prune the less important ones. The final step is to retrain the network to learn the final weights. This step is critical because it may affect the accuracy if pruned network is used without retraining. However it is time consuming, especially in larger networks.

The need of lightweight networks for pose estimation became a necessity to implement HPE in resource constrained environments and embedded systems and to obtain real time performance without much latency. Researchers have developed lightweight structures replacing the standard convolutions of state-of-the-art models like Hourglass and HRNet. Kim and Lee [2020] proposed a multidilated lightweight stacked hourglass in which they replaced standard convolution with depthwise separable blocks which helped to reduce the number of parameters and added dilation to increase the receptive field without adding more parameters. Since depthwise separable convolution separates spatial and channelwise convolutions there can be limited interaction between channel and spatial information which may further lead to loss in representational power.

The introduction of transformers and attention mechanisms created a significant advancement in pose estimation by capturing spatial and temporal dependencies. Vaswani et al. [2017] introduced a transformer which depends on self-attention mechanism which computes the relationship of each element with other elements which helps the model to capture long range dependencies. Even though transformers are originally developed for natural language processing tasks, they have made a remarkable contribution in computer vision tasks because of their capability to model long range dependencies. Unlike traditional CNNs which process information locally, self-attention which is the heart of transformers provides a global context for each joint in pose estimation. This information is crucial and helps in detecting complex poses. It helps the model to better understand and map each keypoints. Multi-head attention used in transformers is an advanced form of self-attention as it applies self-attention multi-ple times in parallel. This enables the model to capture plenty of relationships and patterns in data. Zheng et al. [2021] introduced PoseFormer, a transformer which captures spatial and temporal dependencies for 3D pose estimation in videos and obtained state-of-the-art performance. Yuan et al. [2021] proposed a new architecture which integrates transformers to high resolution network (HRNet). Since HRNet preserves high resolution representations throughout the entire network, when it is combined with transformer this approach enhanced the accuracy by preserving detailed spatial features and dependencies. The transformer-based models showed good results. However, they face shortcomings due to increased computational costs making them unsuitable for edge devices. One of the early efforts to make attention lightweight was the Squeeze-and-Excitation (SE) block, introduced by Hu et al. [2018]. The squeeze operation in a SE block captures global spatial information and then the excitation stage takes the squeezed information and learns channel wise dependencies. CBAM [Woo *et al.*, 2018] is another effective and lightweight attention mechanism which pays attention to important channels and spatial regions within the feature maps, hence improving the overall performance. Integration of these models to the neural network will enhance the representational power of the model.

The choice of activation function is an important factor in DNNs. Clevert et al. [2016] introduced ELU, which produces negative outputs for negative inputs to push activations closer to zero. Thus ELU reduces the bias shift effect present in ReLU, which slows down learning and convergence in it. This helps to make training faster and stable. Their experiments showed that ELU outperforms the famous ReLU on multiple benchmark datasets. Taking this into consideration and inspired by the work of Shen et al. [2024], to improve the training speed and to enhance the cross channel interaction between layers, we used the ELU activation function.

## 3 Proposed Methodology

### 3.1 Proposed Lightweight Network Architecture

We use a stacked hourglass network for pose estimation which is a well known deep learning architecture for feature extraction. Keypoints can be localized in a better way because the network combines features from different resolutions. The network has an encoder decoder like structure with a series of downsampling layers followed by upsampling layers with intermediate supervision. By reducing the spatial resolution, the downsampling layers capture global context which helps the network to learn high level features across the image while the upsampling layers concentrate on reconstructing the original spatial resolution to refine the features which was reduced during the downsampling. Skip connections are implemented in the network by directly connecting each downsampling layer to the corresponding upsampling layer at same level. This will allow the model to combine both high level and low-level features. The spatial resolution of the image is reduced as it is passed through the downsampling layers which in turn will lead to loss of some details. To mitigate these issues and to make the network lightweight and perform well we propose to integrate an attention mechanism into the core of the network to assist the network to focus on important features.

Figure 1 represents the overall architecture of the proposed lightweight hourglass model and Figure 2 details the design of the hourglass block. The goal of the lightweight network is to reduce the model size and the computational complexity while maintaining a good accuracy in estimating poses. The original hourglass module consists of residual blocks based on the standard bottleneck block in ResNet [He *et al.*, 2016] with a series of three convolutional blocks together with batch normalisation and an activation function, typically ReLU and a skip connection as shown in Figure 3a. This structure helps to resolve the issue of the vanishing gradient problem and helps gradients to flow smoothly even in deep networks with many layers. Model pruning was one of the traditional methods used to reduce network's size and computational cost [Li *et al.*, 2017]. Another technique to compress the network is quantization which uses the technique of reducing the precision of activations and weights in network to lower bit formats [Wu *et al.*, 2016].

Some lighter networks such as MobileNet [Howard *et al.*, 2017] and its variants are designed specifically for mobile and embedded systems. The concept of depthwise separable convolution is introduced in original MobileNet architecture which helps in reducing the computation and model size. Other subsequent versions, MobilenetV2 [Sandler *et al.*, 2018] and MobileNetV3 [Howard *et al.*, 2019] introduced inverted residual blocks and linear bottlenecks in order to further optimize the network. In our proposed model we use the concept of depthwise separable convolution in the residual blocks of the Hourglass module. We propose LAP which uses a depthwise separable convolution instead of standard $3 \times 3$ convolution in the bottleneck block of residual unit. Furthermore, we introduce a CBAM to the hourglass module to make the network focus on important features selectively without adding significant computational load.

### 3.2 Depthwise Separable Convolution

Depthwise separable convolution separates standard convolution process into two steps which is depthwise and pointwise. In the standard convolution every filter is applied across all the input channels which makes it computationally expensive in deep networks which have many channels and filters. If an input has M channels, then the depthwise convolution uses M filters which treats each channel independently and does not mix information across channels. Afterwards, a pointwise convolution ($1\times1$ convolution) is applied to the resulting feature map to recombine information from all the channels. We propose LAP which uses depthwise separable convolution instead of standard one in the bottleneck residual

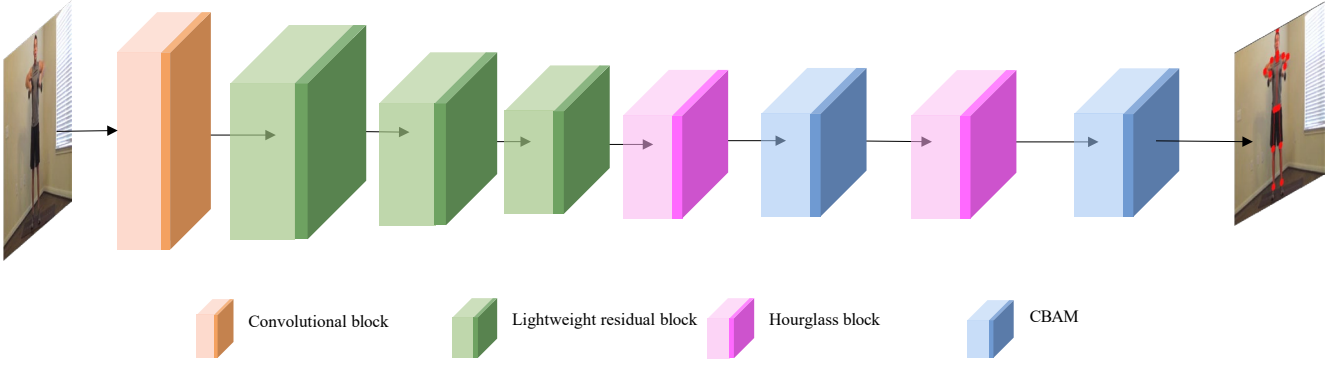

Figure 1: Overall architecture of LAP. The model comprises a convolutional block (orange), lightweight residual blocks (green), hourglass modules (pink), and CBAM blocks (blue).

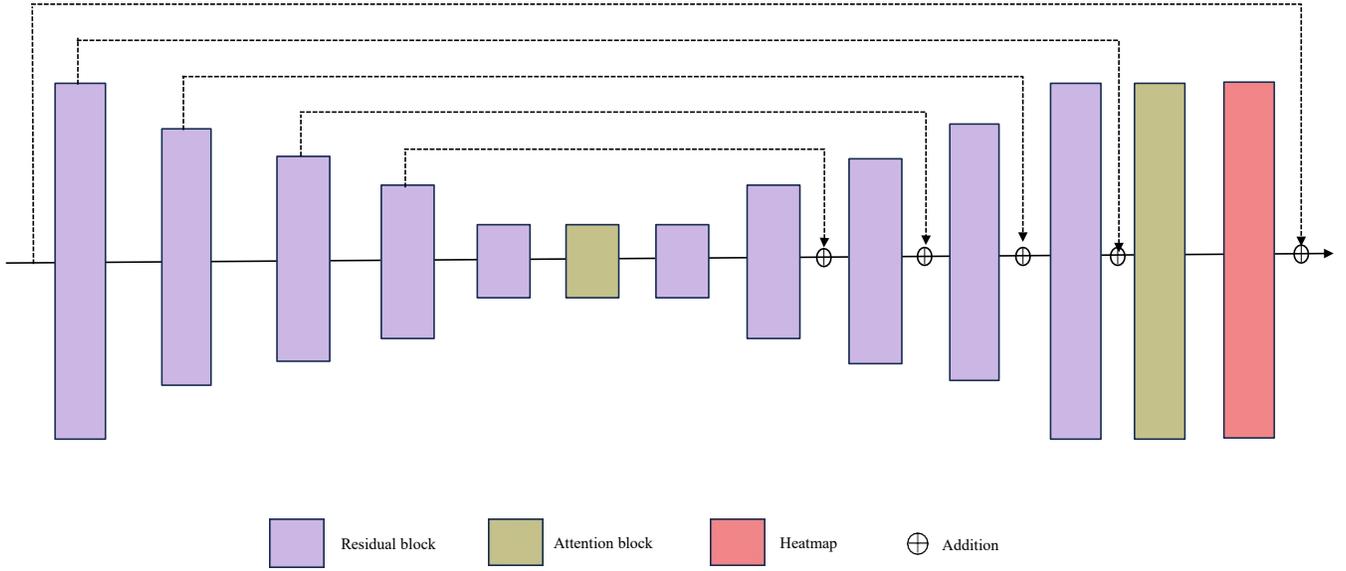

Figure 2: Architecture of the proposed lightweight hourglass block with integrated attention module. The full architecture consists of two such hourglass blocks, each containing residual blocks (purple), attention blocks (CBAM (green)) and generated heatmap (red). The layers are connected with skip connections to enhance feature propagation.

block as shown in Figure 3b.

If $C_{in}$ is the number of input channels and $C_{out}$ is the number of output channels, and the convolution filter size is $D_k \times D_k$, then the number of parameters, $N_S$ of a standard convolution is:

$$N_S = D_k \times D_k \times C_{in} \times C_{out}.$$

The depthwise separable convolution applies separate filters to each input channel individually. Therefore, the number of parameters, $N_D$ for a depthwise separable convolution will be:

$$N_D = D_k \times D_k \times C_{in} + 1 \times 1 \times C_{in} \times C_{out}.$$

The number of floating point operations, $F_S$ in a standard convolution is:

$$F_S = K \times K \times C_{in} \times C_{out} \times D_k \times D_k,$$

where $K \times K$ represents the spatial dimensions (assuming the feature map is square). In the case of a depthwise separable convolution, it acts on a single channel at a time. So for each channel, the FLOPs, $F_{single}$ will be:

$$F_{single} = K \times K \times 1 \times D_k \times D_k,$$

and this sums to $F_{sum}$:

$$F_{sum} = K \times K \times C_{in} \times D_k \times D_k.$$

The pointwise convolution uses $1 \times 1$ kernels to mix information across the channels, and the number of FLOPs for a

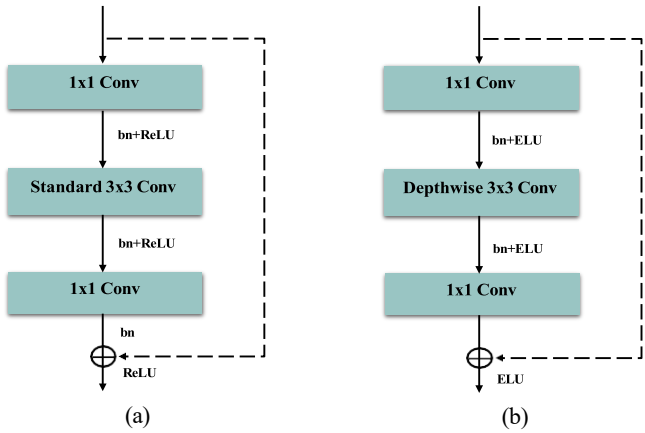

Figure 3: Architecture of bottleneck residual blocks. (a) Standard bottleneck block in ResNet. (b) The lightweight bottleneck block with depthwise separable convolution.

pointwise convolution, $F_p$:

$$F_p = K \times K \times C_{in} \times C_{out}.$$

The total FLOPs in a depthwise separable convolution, $F_D$ are:

$$F_D = K^2 \times C_{in} \times D_k^2 + K^2 \times C_{in} \times C_{out}.$$

We can factor out the common terms. So, the simplified expression is:

$$F_D = K^2 \times C_{in} \times (C_{out} + D_k^2).$$

### 3.3 Attention Mechanism

To improve the performance of our lightweight network, we incorporate attention mechanism using CBAM block. The attention mechanism helps to address the "what" (what features across the channels are important handled by channel attention) and "where" (where in the feature map the critical information is located handled by spatial attention) aspects of feature representations. CBAM consists of two modules which are the Channel Attention Module (CAM) and the Spatial Attention Module (SAM). In Channel Attention, global average pooling and global max pooling aggregate spatial information across each channel separately. These features are then fed to a shared multi-layer perceptron (MLP). The processed output is then passed through a sigmoid activation function for weight normalization. Formally, the channel attention map $M_c(\cdot)$ of a feature map F is computed as,

$$M_c(F) = \sigma\left(\text{MLP}(\text{AvgPool}(F)) + \text{MLP}(\text{MaxPool}(F))\right),$$

where $\sigma(\cdot)$ is the sigmoid activation function. Next, the channel attention map is multiplied with the original feature map to produce the refined feature map:

$$F' = M_c(F) \odot F,$$

where $\odot$ denotes element wise multiplication. To enhance spatial features, spatial attention is performed on the feature map. Across the channel dimension, we perform global average pooling and global max pooling and then a $7 \times 7$ convolution is performed on the obtained feature maps. The resulting spatial attention map is then multiplied with the original feature map.

The spatial attention map, $M_s(\cdot)$ is calculated as,

$$M_s(F) = \sigma\left(f^{7\times 7}([\text{AvgPool}(F); \text{MaxPool}(F)])\right),$$

where $f^{7\times 7}(\cdot)$ denotes the convolution operation with a $7 \times 7$ kernel.

The spatial attention map is then applied to the feature map refined by the channel attention and the final feature map is:

$$F'' = M_s(F') \odot F'.$$

The output from the channel attention module is fed as an input (broadcasted) to the spatial attention module. The network can more effectively utilize both inter-channel and intra-spatial information from this sequential implementation.

## 4 Experiments
### 4.1 Datasets

Famous pose estimation dataset COCO 2017 [Lin *et al.*, 2014] with its keypoint annotations were primarily used for training and evaluating the model. Additionally, the MPII [Andriluka *et al.*, 2014] with its keypoint annotations were also used to train the model. The COCO dataset which is widely used for tasks such as segmentation, human pose estimation and object detection contains more than 200,000 images which is divided into train, validation and test sets. It has 250,000 individual keypoint instances with 17 keypoints per person. The 17 keypoints include nose, left eye, right eye, left ear, right ear, left shoulder, right shoulder, left elbow, right elbow, left wrist, right wrist, left hip, right hip, left knee, right knee, left ankle, right ankle. Our model is trained on the COCO train 2017 dataset and validated on the COCO validation 2017 dataset. The MPII dataset consists of around 25,000 images in which people perform a wide range of activities. It contains more than 40,000 person instances extracted from YouTube. Each person instance is annotated with 16 keypoints including head, neck, pelvis, thorax, left shoulder, right shoulder, left elbow, right elbow, left wrist, right wrist, left knee, right knee, left ankle, right ankle, left hip and right hip. Here nose, eyes and ears are not explicitly annotated as in the COCO dataset.

To enhance the diversity of the dataset and to improve the model's generalization we employed several data augmentation techniques. Techniques such as random scaling ($\pm 1.0$ to $\pm 3.0$), random rotation and random flipping were applied. Random color jittering is applied to randomly alter the brightness and contrast of the image which helps the model to address varying lighting conditions. We also performed crop-ping around the region of interest (person). These augmentations together contribute to the model's ability to generalize unseen data in a better manner.

### 4.2 Training Details

The research was conducted on the Katana [UNSW, 2021] computational cluster, supported by Research Technology Services at UNSW Sydney. The experiment was conducted on Linux kernel utilizing NVIDIA Tesla V100-SXM2-32GB GPU. The implementation was based on the PyTorch framework, version 2.3.0 and CUDA 11.8 with Python version 3.10.8. The optimizer used was Adam with a batch size of eight. The model was trained for 40 epochs and the initial learning rate was set to $2.5 \times 10^{-4}$. We utilized ReduceLROnPlateau learning rate scheduler to reduce the learning rate by a factor of 0.2, if there is no improvement in loss continuously for five epochs. This is done to enhance the model's stability during training. The COCO dataset was processed with an image size of $256 \times 192$, while the MPII dataset used an image size of $256 \times 256$. See Table 1 for hyperparameter settings.

Table 1: Hyperparameter Settings

| Parameter | Settings |
| --- | --- |
| Optimizer | Adam |
| Batch size | 8 |
| Learning rate | $2.5 \times 10^{-4}$ |
| Epoch | 40 |
| Learning rate schedule | ReduceLROnPlateau |
| Patience | 5 epochs |
| Image size (COCO) | $256 \times 192$ |
| Image size (MPII) | $256 \times 256$ |

### 4.3 Loss Calculation and Evaluation Metrics

The Mean Squared Error (MSE) loss is a commonly used loss function for pose estimation tasks. The MSE loss calculates the difference between predicted heatmap values and ground truth heatmap values. In pose estimation tasks each keypoint is represented as a heatmap (Gaussian distribution with a peak representing the most likely position of a keypoint at that location). The ground truth heatmap for each keypoint is constructed using a Gaussian peak of the keypoint coordinates. During training the model predicts its own heatmaps aiming to position Gaussian peaks at correct keypoint positions. MSE loss is then calculated as the average of the squared differences between the predicted heatmaps and the ground truth heatmaps across all keypoints and all pixels in the heatmap. This function measures the deviation of predicted keypoint locations from that of the ground truth. The model uses this information to adjust the parameters to reduce loss in subsequent iterations. MSE is defined as:

$$\text{MSE} = \frac{1}{N} \sum_{i=1}^{N} (H_i - \hat{H}_i)^2, \qquad (1)$$

where $N$ is the number of samples in the batch, $H_i$ is the ground truth heatmap value at position (x,y) for the $i$-th sample, $\hat{H}_i$ is the predicted heatmap value at position (x,y) for the $i$-th sample.

Figure 4 illustrates the loss curves for both the COCO and MPII datasets across multiple epochs. The loss curves are plotted to monitor the model's convergence during training. As illustrated in Figure 4a the training loss and validation loss are decreasing and after the 30th epoch they decrease but with slight fluctuations. Figure 4b illustrates a similar trend for the COCO dataset where losses decrease as training progresses. However the COCO dataset exhibits a smaller gap between losses suggesting that the model is less prone to overfitting.

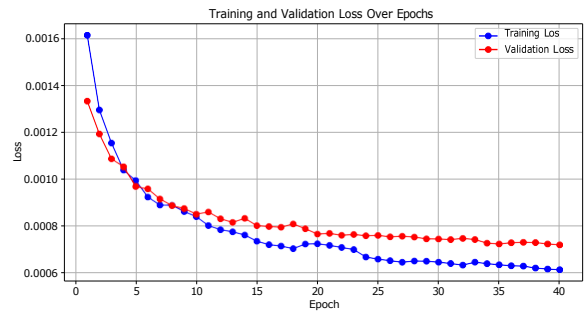

(a) Training and validation loss over epochs for the MPII dataset.

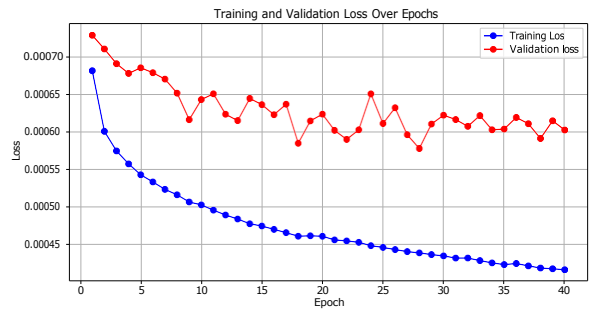

(b) Training and validation loss over epochs for the COCO dataset.

Figure 4: Training and validation loss plotted against the number of epochs for both MPII and COCO datasets over 40 epochs.

The performance of the model is evaluated using metrics such as Average Precision (AP). AP is a widely used metric for pose estimation and object detection which is calculated by measuring the area under the precision-recall curve for different threshold values.

Table 2: Comparison with other lightweight models based on the number of parameters, FLOPs, and Average Precision (AP) using COCO 2017 validation dataset (Note: [xStack] represents the number of stacked blocks in model architecture).

| Model Name[xStack] | Year | Backbone | No. of Parameters | FLOPs | AP (%) |
|---|---|---|---|---|---|
| LPN-50 [Zhang *et al.*, 2020] | 2020 | ResNet-50 | 2.90M | 1.0G | 69.10 |
| LPN-101 [Zhang *et al.*, 2020] | 2020 | ResNet-101 | 5.30M | 1.4G | 70.40 |
| LPN-152 [Zhang *et al.*, 2020] | 2020 | ResNet-152 | 7.40M | 1.8G | 71.0 |
| EfficientPose-C [Zhang *et al.*, 2020] | 2020 | MobileNetV2 | 5.0M | 1.6G | 71.30 |
| KD-S-HRNet[Cao *et al.*, 2023] | 2023 | HRNet | 6.22M | 2.15G | 71.70 |
| SWBPose [Fan *et al.*, 2023] | 2023 | Swin Transformer | 4.37M | **0.61G** | 66.01 |
| Hourglass[x2] | 2016 | Hourglass | 6.70M | 9.08G | 71.49 |
| **LAP[x2] (Our method)** | **2024** | **Hourglass** | **2.30M** | 3.7G | **72.07** |

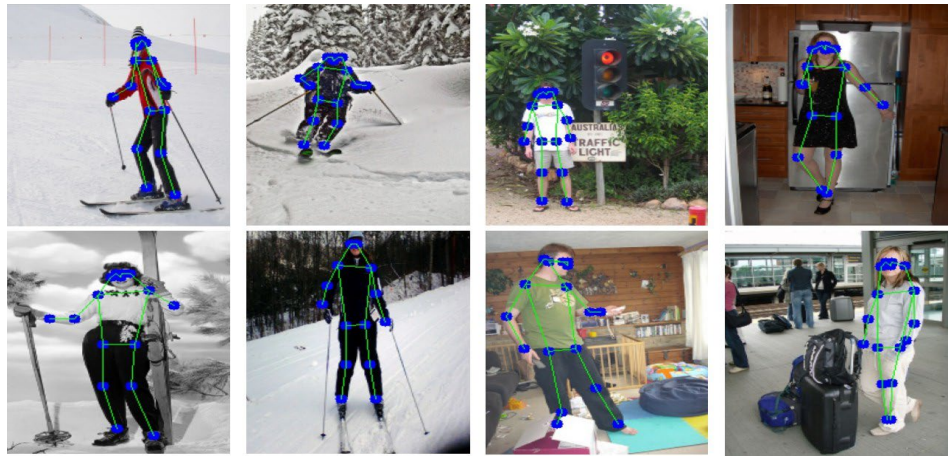

Figure 5: Pose estimation results from model trained with COCO dataset. The results shows the identification of keypoints across different poses.

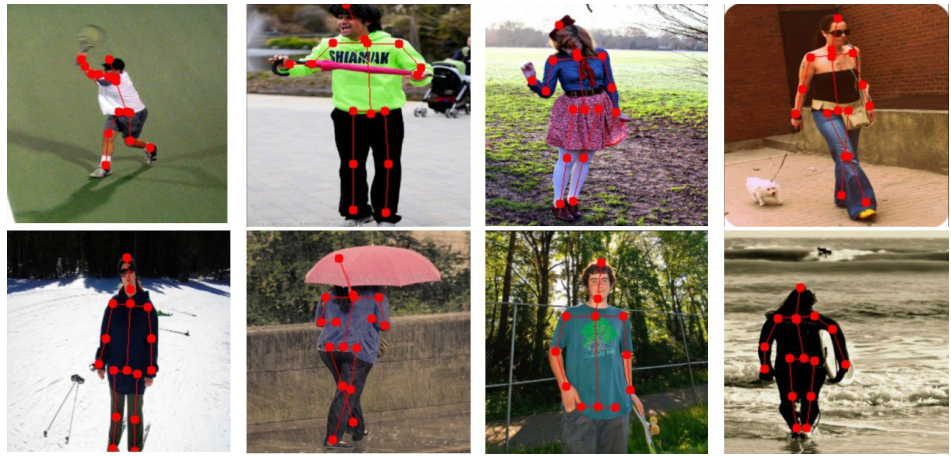

Figure 6: Pose estimation results from model trained with MPII dataset. The results shows the identification of keypoints across different poses.

## 5 Result Analysis

Table 2 presents the comparison of the proposed model with some other human pose estimation models on COCO 2017 validation dataset. These models were chosen for comparison because they use state-of-the-art networks such as ResNet, MobileNet, HRNet and Swin Transformer as their backbone. Figure 5 and 6 visualizes the pose estimation results from the model when trained with COCO dataset and MPII dataset respectively. Figure 7 represents the AP as a function of the number of epochs. The values of AP, FLOPs and number of parameters of other lightweight networks used for comparison were taken from respective sources as cited in Table 2. The authors conducted their experiments on same dataset (COCO), so we use the values reported from their results in our comparison. We implemented, state-of-the-art hourglass with 2 stacks. The proposed model achieves an AP of 72.07%, outperforming other lightweight models mentioned in the Table 2. Zhang et al. [2020] proposed Lightweight Pose Network (LPN), a lightweight pose estimation model using different ResNet backbones. While SWB-Pose achieves lower FLOPs compared to our model, its AP is 66%, which is lower than the AP of our model. LAP have fewer parameters when compared with other models. Overall, LAP maintains a balance between computational efficiency and accuracy, while achieving higher AP value. Our proposed method is approximately 65.67% smaller in terms of parameters and uses around 59.27% fewer FLOPs, while still achieving good performance compared to state-of-the-art hourglass networks when implemented using 2 stacks.

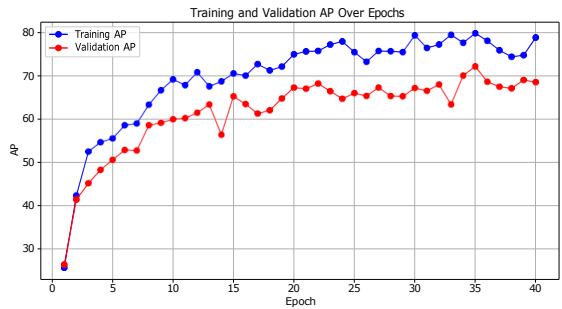

Figure 7: Training and validation AP over epochs for the COCO dataset.

## 6 Conclusion and Future Works

This paper proposes a lightweight human pose estimation based on hourglass and convolutional block attention module. The model contains fewer parameters and less computational cost which makes it a plausible lightweight alternative. The integrated attention module helps the model become more effective in detecting human keypoints by prioritizing important regions in the heatmap while suppressing the less important ones. The lightweight residual block with depthwise separable convolution takes part in reducing the number of parameters and the FLOPs of the model. Experiments were performed on COCO and MPII datasets. The number of parameters and computational complexity of network is substantially reduced with minor performance loss as compared to state-of-the-art methods. Compared to the dual stack hourglass model, which has 6.7 million parameters, our model significantly reduces the parameter count to 2.3 million along with fewer FLOPs. The results prove that LAP can effectively balance prediction accuracy with complexity, making it suitable for pose estimation in resource constrained applications. However, some underexplored challenges in pose estimation includes occlusions and uneven lighting scenarios. Joints can be hard to detect in case of occlusions. For instance, self occlusions and occlusions caused by another person or object or the clothing patterns of person and human body structure are critical areas. Future work of this research will focus on exploring and addressing these challenges by incorporating various occlusion handling techniques, such as bio-mechanical modelling.